\documentclass[conference]{IEEEtran}
\IEEEoverridecommandlockouts
\usepackage{cite}
\usepackage{amsmath,amssymb,amsfonts}
\usepackage{algorithmic}
\usepackage{graphicx}
\usepackage{textcomp}
\usepackage{xcolor}

\usepackage{multicol}
\usepackage{graphicx}
\usepackage{textcomp}
\usepackage{xcolor}
\usepackage{microtype}
\usepackage{graphicx}
\usepackage{subcaption}

\usepackage{booktabs} 
\usepackage{hyperref}
\usepackage{caption}
\usepackage{cleveref}
\usepackage{makecell}
\usepackage{dsfont}
\usepackage{multirow}
\usepackage{svg}
\usepackage{cleveref}
\usepackage{bbm}
\usepackage{pifont}
\usepackage{pythonhighlight}
\usepackage{listings}
\usepackage{graphicx} 
\usepackage{amsmath}
\usepackage{optidef}
\usepackage{float}
\usepackage{epsfig}
\usepackage{svg}
\usepackage{pdflscape}
\usepackage{rotating}

\lstdefinestyle{Python}{
    language        = Python,
    basicstyle      = \ttfamily,
    keywordstyle    = \color{blue},
    keywordstyle    = [2] \color{teal}, 
    stringstyle     = \color{green},
    commentstyle    = \color{red}\ttfamily
}

\definecolor{codegreen}{rgb}{0,0.6,0}
\definecolor{codegray}{rgb}{0.5,0.5,0.5}
\definecolor{codepurple}{rgb}{0.58,0,0.82}
\definecolor{backcolour}{rgb}{0.95,0.95,0.92}

\def\BibTeX{{\rm B\kern-.05em{\sc i\kern-.025em b}\kern-.08em
    T\kern-.1667em\lower.7ex\hbox{E}\kern-.125emX}}
\begin{document}
\title{XTSC-Bench: Quantitative Benchmarking for Explainers on Time Series Classification \thanks{This work was carried out with the support of the German Federal Ministry of Education and Research (BMBF) within the project "MetaLearn" (Grant 02P20A013).}}
\author{
\IEEEauthorblockN{ Jacqueline Höllig\IEEEauthorrefmark{1},
Steffen Thoma\IEEEauthorrefmark{1}, Florian Grimm\IEEEauthorrefmark{2}}
\IEEEauthorblockA{\IEEEauthorrefmark{1} FZI Research Center for Information Technology, Karlsruhe, Germany,
\{hoellig, thoma\}@fzi.de}%
\IEEEauthorblockA{\IEEEauthorrefmark{2} ESB Business School, Reutlingen, Germany, Florian.Grimm@Reutlingen-University.de}%
}
\maketitle


\begin{abstract}
Despite the growing body of work on explainable machine learning in time series classification (TSC), it remains unclear how to evaluate different explainability methods. Resorting to qualitative assessment and user studies to evaluate explainers for TSC is difficult since humans have difficulties understanding the underlying information contained in time series data. Therefore, a systematic review and quantitative comparison of explanation methods to confirm their correctness becomes crucial. 
While steps to standardized evaluations were taken for tabular, image, and textual data, benchmarking explainability methods on time series is challenging due to a) traditional metrics not being directly applicable, b) implementation and adaption of traditional metrics for time series in the literature vary, and c) varying baseline implementations.
This paper proposes XTSC-Bench, a benchmarking tool providing standardized datasets, models, and metrics for evaluating explanation methods on TSC. We analyze 3 perturbation-, 6 gradient- and 2 example-based explanation methods to TSC showing that improvements in the explainers' robustness and reliability are necessary, especially for multivariate data. 
\end{abstract}


\section{Introduction}
As the use of machine learning models, especially deep learning, increases in various domains ranging from health care \cite{markus2021role} to predictive maintenance  \cite{vollert2021interpretable}, the need for reliable model explanations is also growing. 
An increasing number of methods providing a variety of explanation types (e.g., example-based methods like counterfactuals \cite{wachter2017counterfactual}, or feature attribution methods like SHAP \cite{lundberg2017unified}) on different data types (e.g., images \cite{smilkov2017smoothgrad}, 
tabular data \cite{ribeiro2016should}) are available. However, measuring the performance of such explanation methods is still a challenge. There are no generally agreed upon-metrics measuring the quality of explanations, and comparisons between different implementations and metrics are difficult (e.g., \cite{carvalho2019machine,lipton2018mythos,schmidt2019quantifying}). 
The first steps to standardize the metrics notion and implementation have been taken by different frameworks implementing explainability algorithms (e.g., Captum \cite{kokhlikyan2020captum}, AIX360 \cite{aix360-sept-2019}) and Quantus \cite{hedstrom2023quantus}, a framework dedicated to the evaluation of explanations. 
The main focus of those frameworks is to provide explainability to image, tabular, and textual classification tasks.
Although time series classification (TSC) is a ubiquitous task, it has been neglected. Due to the different structure and properties of time-ordered data, the application of non-time-specific explanation algorithms is not advisable, leading to a new subfield in Explainable Artificial Intelligence (XAI) - Explainable Time Series Classification (XTSC) \cite{rojat2021explainable}.
\par
While the first step to standardize the explanation benchmarking process for the time series domain has been taken by TSInterpret \cite{hollig2023tsinterpret} - a framework implementing explanation methods for time series classification in a unified interface -  standardized metrics for evaluating the quality of explanation methods are still missing \cite{theissler2022explainable}. 
Similar to the explanation methods implemented in the different explainability frameworks, transferring metrics to the time series domain is complex. 
Using metrics from traditional frameworks (e.g., \cite{hedstrom2023quantus}) can lead to erroneous assumptions in the time series domain. 
This lack of specific and standardized metric and baseline implementations lead to a high variety of proposed metrics, metric implementations, proposed baselines, and baseline implementations.
\par
In this paper, we propose XTSC-Bench, a benchmarking tool implementing a variety of metrics for a standardized and systematic evaluation of explainers for TSC. 
Its connection to TSInterpret \cite{hollig2023tsinterpret} ensures a unified implementation of benchmarking algorithms. We utilize XTSC-Bench to evaluate 3 gradient- and 6 perturbation-based feature importance methods and 2 example-based approaches.
Our contribution is twofold: 
\begin{itemize}
    \item A thorough investigation of existing approaches.
    \item An easy-to-use benchmarking tool compatible with TSInterpret 
    \cite{hollig2023tsinterpret}.
\end{itemize}
\section{Related Work}\label{sec:RelatedWork}
\begin{table*}[t]
 \centering
 \tiny
 \caption{Evaluation settings of explainers for TSC. Bold are the explanation algorithms evaluated in \Cref{sec:Emp}. If no metric source is provided, the paper authors did not specify which notation was used. The * indicates that only a subset of the dataset was used.}
\label{tab:EvalTimeSeries}

\begin{tabular}{|c|c|c|c|c|c|}
\hline
    \textbf{Explainer} & \textbf{Dataset} &  \textbf{Metrics / Target}& \textbf{Baselines}& \textbf{TS-Baselines}\\
    \hline
     \textbf{LEFTIST} \cite{guilleme2019agnostic} & UCR Archive \cite{dau2019ucr}* &Faithfulness, Understandable \cite{guilleme2019agnostic} & - & -\\
    \hline
    
    \textbf{NG} \cite{delaney2021instance}& UCR Archive \cite{dau2019ucr}* & Proximity, Sparsity, Plausibility, Diversity \cite{pawelczyk2021carla} & W-CF \cite{wachter2017counterfactual} & NUN-CF\cite{delaney2021instance}\\ 
    \hline
    \textbf{TSEvo} \cite{hollig2022tsevo} & UCR Archive \cite{dau2019ucr}*&  Proximity, Sparsity, Plausibility \cite{pawelczyk2021carla}& W-CF \cite{wachter2017counterfactual} & NG \cite{delaney2021instance}, COMTE \cite{ates2021counterfactual}\\
    \hline
    \textbf{TSR} \cite{ismail2020benchmarking} &Synthetic Data \cite{ismail2020benchmarking} & Reliability \cite{ismail2020benchmarking} & \makecell{GradCam \cite{selvaraju2017grad}, \\ Integrated Gradients \cite{sundararajan2017axiomatic},\\ Feature Occlusion\cite{zeiler2014visualizing} \& more} & -\\ 
    \hline
   
    COMTE\cite{ates2021counterfactual} & \makecell{Hpas, Taxonomist,\\ Cori, NATOPS}& \makecell{ Complexity \cite{miller2019explanation}, Faithfulness\cite{ribeiro2016should}, Robustness \cite{alvarez2018towards}}& SHAP \cite{lundberg2017unified}, Lime \cite{ribeiro2016should}& -\\
    \hline
    LASTS \cite{guidotti2020explaining}& UCR/UEA Archive & Faithfullness \cite{guidotti2018survey,freitas2014comprehensible}, Robustness& SHAP \cite{lundberg2017unified} & SHAPS \cite{schlegel2019towards}\\
    \hline
    SETS \cite{bahri2022shapelet}&Solar Flare Dataset&Proximity, Sparsity, Plausibility \cite{pawelczyk2021carla}&& NG \cite{delaney2021instance}, COMTE \cite{ates2021counterfactual},\\ 
    \hline
    TSInsight \cite{siddiqui2021tsinsight} & UEA Dataset \cite{bagnall2018uea}* & Faithfullness \cite{fong2019understanding}& \makecell{GradCam \cite{selvaraju2017grad}, Gradient x Input \cite{shrikumar2016not},\\ Feature Occlusion \cite{zeiler2014visualizing} \& more}  &-\\
    \hline
    TSViz \cite{siddiqui2019tsviz} && Complexity, Faithfullness, Robustness \cite{siddiqui2019tsviz}&-&-\\ 
\hline
\end{tabular}
\end{table*}
Several researchers stress the need for formal evaluation metrics and a more systematic evaluation of explainability methods \cite{guidotti2018survey} \cite{theissler2022explainable}. 
For image, tabular, and textual data, a standard is slowly emerging \cite{bhatt2020evaluating} \cite{nguyen2020quantitative} with easy-to-adopt frameworks and the inclusion of some general metrics into explanation frameworks (e.g., \cite{aix360-sept-2019}, \cite{kokhlikyan2020captum}) as well as a framework dedicated to quantization \cite{hedstrom2023quantus}. Nonetheless, due to the relative newness of explainability to Deep Learning for TSC\footnote{First XAI approaches on time series were only emerging in the last decade. For a detailed survey, we refer the reader to \cite{rojat2021explainable}.}, standardization for benchmarking explainability algorithms on time series is still missing. \Cref{{tab:EvalTimeSeries}} shows the evaluation settings of various explanation algorithms for TSC. While the data basis is mostly standardized, i.e., most algorithms use a subset of data included in the UCR \cite{dau2019ucr} or UEA Archive \cite{bagnall2018uea}, all of the algorithms included in \Cref{tab:EvalTimeSeries} rely on comparing the newly developed algorithm with a time series unspecific algorithm. However, it has been shown that time-series unspecific explanation algorithms are not able to capture the time component sufficiently as they rely heavily on independent feature assumption and cannot uncouple the feature and time domain \cite{ismail2020benchmarking}. 
Although most metrics used in these evaluations have the same evaluation target, i.e., faithfulness, robustness, and  reliability\footnote{Proximity, sparsity, diversity and plausibility are counterfactual-specific evaluation metrics and therefore not applicable to all explainer types.}, their definitions and implementations differ. Often, metrics are directly transferred from image classification \cite{siddiqui2021tsinsight}. However, many metrics rely on replacing input parts with uninformative information (e.g., to measure if the explanation method shows the same behavior as the classifier) or on comparisons to segmentation masks (e.g., to measure if the explanation method was able to localize relevant features).  While providing uninformative features is trivial for e.g., images by replacing parts of an image with black or white pixels \cite{bhatt2020evaluating}, replacing features with standard techniques (class means or zeros) might be relevant information in time series. 

\section{Problem Definition}\label{sec:Problem}
We study a supervised TSC problem. Let $x = [x_{11}, . . . , x_{NT}] \in {\rm I\!R}^{N \times T}$ be a uni- or multivariate time series, where $T$ is the number of time steps, and $N$ is the number of features. Let $x_{i,t}$ be the input feature $i$ at time $t$. Similarly, let $X_{:,t} \in  {\rm I\!R}^N$ and $X_{i,:} \in  {\rm I\!R}^T$ be the feature vector at time $t$, and the time vector for feature $i$, respectively. $Y$ denotes the output, and $f:x \rightarrow Y$ is a classification model returning a probability distribution vector over classes $Y= [y_{1}, ..., y_{C}]$, where $C$ is the total number of classes (i.e., outputs) and $y_{i}$ the probability of $x$ belonging to class $i$. An explanation method $E_{f}$ finds  an explanation $E_{f}(X) \in  {\rm I\!R}^{N \times T}$. In the case of feature attribution methods, the explainer  $E_{f}$ assigns an attribution $a_{it}$ to explain the importance of a feature $i$ a time step $t$, resulting in $E_{f}(X)=(a_{11},...,a_{NT})$. 
For example-based methods $E_{f}$ provides an example with the same prediction or a counterexample resulting in $E_{f}(X)=(x'_{11},...,x'_{NT})$.
\par
\begin{figure}[tb]
     \centering
     \begin{subfigure}{0.4\columnwidth}
         \centering
         \includegraphics [width=0.9\textwidth]{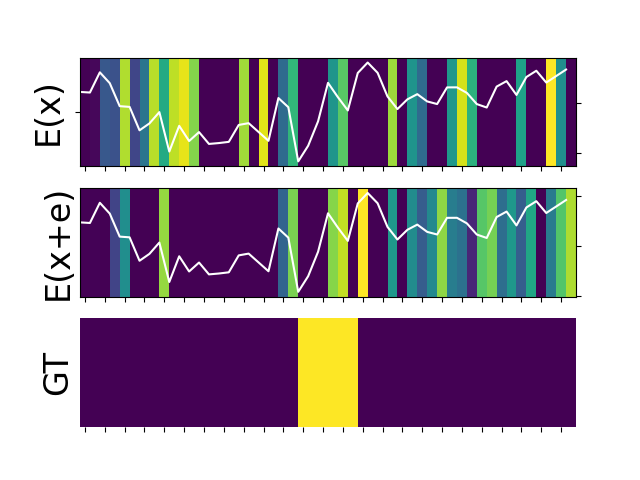}
         \caption{Gradient-Based $E$}
         \label{fig:grad}
     \end{subfigure}
     \hfill
     \begin{subfigure}{0.4\columnwidth}
         \centering
         \includegraphics [width=0.9\textwidth]{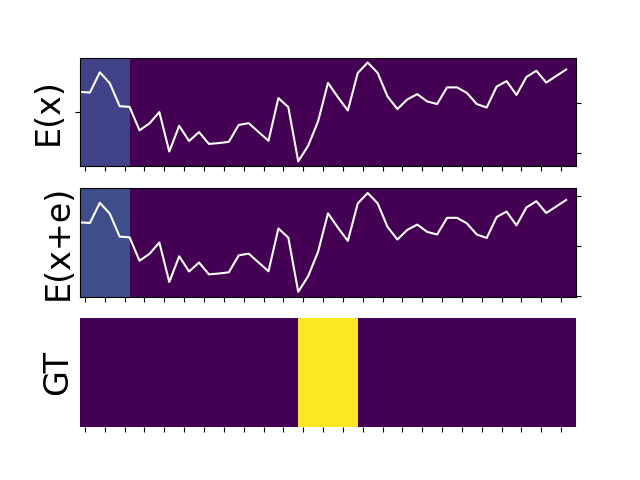}
         \caption{Perturbation-Based $E$}
         \label{fig:leftist}
     \end{subfigure}
\caption{Visualization of metric implications on a sample explanation $E(x)$.
}
\label{fig:MetricImplication}
\end{figure}

For an Explainer $E_{f}$ to provide good explanations, those explanations need to be: 
\begin{itemize}
    \item Reliable
    : An explanation should be centered around the region of interest, the ground truth $GT$. 
    \begin{equation*}
    \scriptsize
        E_{f}(x) \cong GT
    \end{equation*}
    \item Faithful
    :  The explanation algorithm $E_{f}$ should replicate the models $f$ behavior. 
    \begin{equation*}
    \scriptsize
    E_{f}(x) \sim f(x)
    \end{equation*}
    \item Robust
    : Similar inputs should result in similar explanations. 
    \begin{equation*}
    \scriptsize
        E_{f}(x) \approx E_{f}(x + \epsilon)
    \end{equation*}
    \item Complex
    : Explanations using a smaller number of features are preferred. It is assumed that explanations using a large number of features are difficult for the user to understand \cite{miller2019explanation}.
    \begin{equation*}
    \scriptsize
        \min \mathds{1}_{E_{f}(x)>0}
    \end{equation*}
    
\end{itemize}
\Cref{fig:MetricImplication} visualizes the implications of the requirements above on explanations obtained from an gradient-based explainer (an explanation based on the classifiers gradient estimations) and a perturbation-based explainer (an explanation based on observing the influence of input modifications). 
The top images show the original time series $x$ with an explanation $E(x)$ visualized as a heatmap. The middle image shows the perturbed time series $x + \epsilon$ with the explanation $E(x+\epsilon)$. The bottom image shows the known ground truth $GT$. 
In case of this specific time series: The complexity is high for \Cref{fig:grad}, resulting from the many attributions (highlights). For \Cref{fig:leftist} the complexity is low. Although  \Cref{fig:leftist} performs better on complexity taking the ground truth $GT$ into account, the attributions obtained on the sample are inconsistent with $GT$. The explanation in \Cref{fig:leftist} is more robust than \Cref{fig:grad} as the explanations $E(x)$ and $E(x + \epsilon)$ are identical. Faithfulness quantifies the consistency between the decision-making process of $f$ and the explanations $E$. The consistency of \Cref{fig:grad} is higher than the one from \Cref{fig:leftist} as \Cref{fig:grad} relies on the gradients of $f$ while \Cref{fig:leftist} fits a surrogate model. Overall, in this case, although \Cref{fig:leftist} performs better on complexity and robustness than \Cref{fig:grad}, due to the limited reliability (i.e., consistency with $GT$), \Cref{fig:grad} should be the preferred explainer.
\section{XTSC-Bench: A Benchmarking  Tool}
\begin{figure}[tb]
    \centering
    \includegraphics [width=0.9\columnwidth]{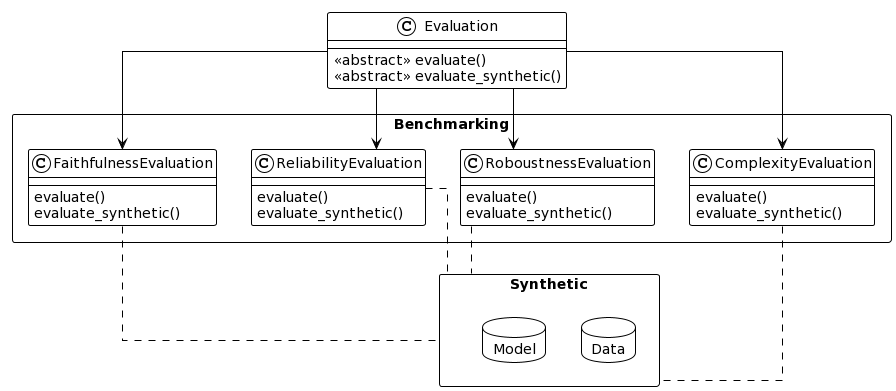}
    \caption{Architecture of XTSC-Bench.}
    \label{fig:arc}
\end{figure}

The goal of XTSC-Bench is to provide a simple and standardized framework to allow users to apply and evaluate different State-of-the-Art explanation models in a standardized and replicable way on the notions of complexity, reliability, robustness, and faithfulness. \Cref{fig:arc} visualizes the architecture. The  benchmarking tool is split according to \Cref{sec:Problem} into different classes for benchmarking reliability, faithfulness robustness, and complexity. As some of the notions (e.g., reliability) rely on a fairly accurate definition of an explanation ground truth $GT$ or the iterative masking of parts of the original input with known uninformative features, we include uni- and multivariate synthetic data and pre-trained models in the benchmarking tool (see \Cref{sec:Synth}). Each class follows the evaluation interface, providing a method \emph{evaluate} and a method \emph{evaluate\_synthetic}. The function \emph{evaluate} allows the usage of non-synthetic data and models as well as the evaluation of a single explanation on-the-fly.  For all metrics we use a wrapper build around Quantus \cite{hedstrom2023quantus} and added some time-specific tweaks.\par 

\subsection{Synthetic Data and Pretrained Models}\label{sec:Synth}
\begin{figure}[tb]
    \centering
   \begin{subfigure}{0.1\textwidth}
   \includegraphics[width=\textwidth]{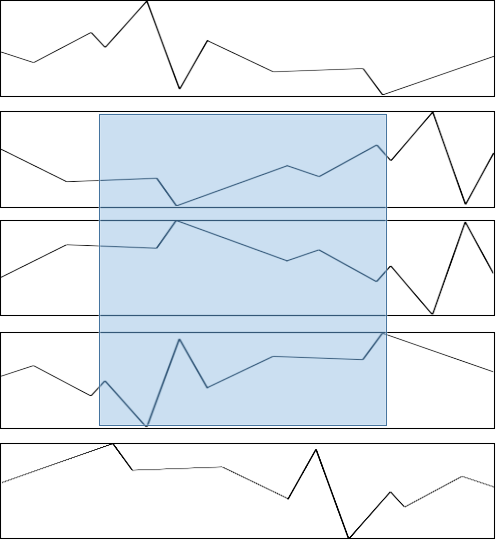} 
   \caption{Box}
   \end{subfigure}
   \begin{subfigure}{0.1\textwidth}
   \includegraphics[width=\textwidth]{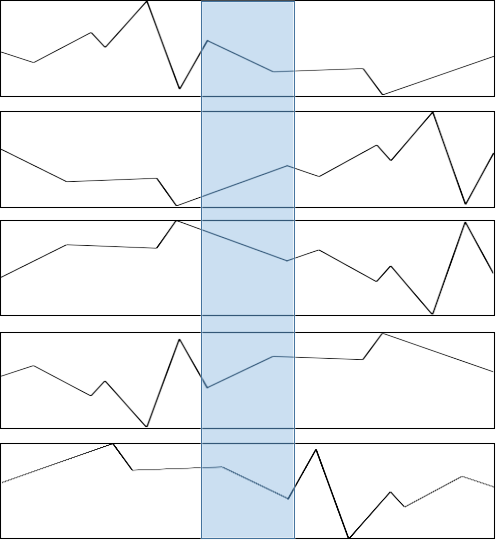} 
   \caption{Rare Time}
   \end{subfigure}
   \begin{subfigure}{0.1\textwidth}
   \includegraphics[width=\textwidth]{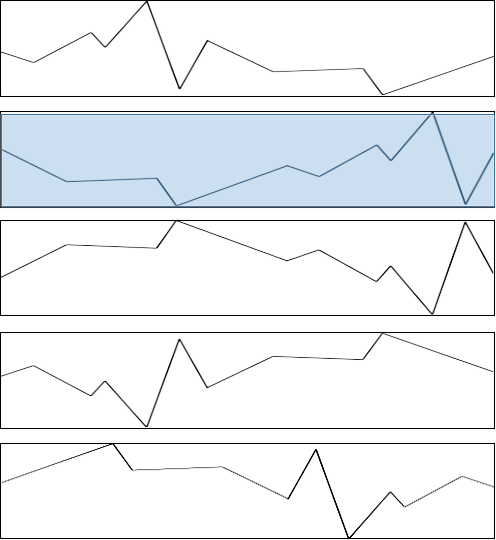}    
   \caption{Rare Feat.}
   \end{subfigure}
   \begin{subfigure}{0.1\textwidth}
   \includegraphics[width=\textwidth]{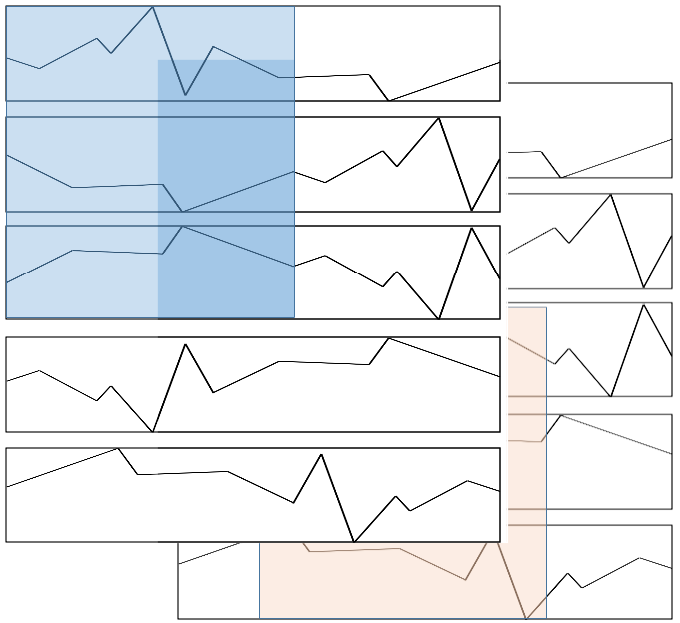} 
   \caption{Moving}
   \end{subfigure}
\caption{Visualization of Informative Features types. The rectangle indicates the informative features.}
\label{fig:InformFeat}
\end{figure}

XTSC-Bench provides 60  uni- and  60 multivariate synthetic datasets with 50 time steps generated according to Ismail et al. \cite{ismail2020benchmarking}\footnote{Find details on the Data Generation in \cite{ismail2020benchmarking} or in the \Cref{sec:ExpApp} in our GitHub Repository: \url{https://github.com/JHoelli/XTSC-Bench/blob/main/Appendix.pdf}.}. The 'base' dataset is  generated based on various time series processes (Gaussian, Autoregressive, Continuous Autoregressive, Gaussian Process, Harmonic, NARMA and Pseudo Periodic). For each 'base' dataset obtained from the time series process, multiple synthetic datasets are obtained by adding various Informative Features ranging from Rare Features (less than 5\% of features) and time steps (less than 5\% of time steps) mimicking an anomaly detection task to boxes covering over 30\% of features and time steps (see \Cref{fig:InformFeat}). A binary label is added for each dataset (time process $\times$ informative feature) by highlighting informative features with the addition of a constant for positive classes and subtraction for negative classes.
For all synthetic uni- and multivariate datasets, we train a 1D-Convolutional Network with ResNet Architecture (CNN) and Long Short Term Memory (LSTM) with a hidden layer of size 10. We train the networks with a cross-entropy loss for 500 epochs with a patience of 20 and Adam with a learning rate of 0.001. The trained networks are also provided in XTSC-Bench.
\subsection{Robustness} \label{sec:Roboust} Robustness measures the stability of an explanation method's output subjected to a slight input perturbation $\bar{x}= x + \epsilon $ under the assumption that the model's output approximately stays the same $f(x) \approx f(\bar{x})$. Small, unmeaningful changes around $x$ should lead to a consistent explanation. XTSC-Bench employs two metrics measuring the robustness of an explanation algorithm $E$:
\begin{itemize}
    \item Max Sensitivity \cite{bhatt2020evaluating} measures the maximum change in the explanation with a small perturbation of the input $x$. $r$ denotes the input neighborhood ratio.
    \begin{equation}
    \scriptsize
        Sens_{max} (E,f,x,r) = max_{\bar{x}-x \le r} ||E_{f}(\Bar{x})-E_{f}(x)||
    \end{equation}
    \item Average Sensitivity \cite{bhatt2020evaluating} denotes the average sensitivity in the neighborhood of $x$ with $\Bar{x}-x \leq r$.
    \begin{equation}
    \scriptsize
        Sens_{mean} (E,f,x,r) = \frac{1}{|x|}\sum ||E_{f}(\Bar{x})-E_{f}(x)||
    \end{equation}

\end{itemize}
\subsection{Faithfulness}\label{sec:Faith}
Faithfulness quantifies the consistency between the prediction model $f$ and explanation model $E$. Most faithfulness metrics rely on so called reference baselines consisting of non-informative features. In literature, those reference baselines are often training data means or zeros (e.g., \cite{sundararajan2017axiomatic}).
However, for time series data those baselines might contain information (e.g., 0 might be an informative anomaly). Therefore, on the proposed synthetic data the reference baseline $\tilde{x}$ is sampled from the generation process. XTSC-Bench employs faithfulness correlation \cite{bhatt2020evaluating} to measure the correlation between the sum of attributions $\sum_{s \in S}E_{f}(x_{x_{s}=\tilde{x}_{s}})$ and the difference in output $f(x)-f(x_{x_{s}=\tilde{x}_{s}})$ when setting those features to a reference baseline $x_{x_{s}=\tilde{x}_{s}}$. $S$ is a subset of input features, $\tilde{x}_{S}$ denotes a subset of the reference baseline $\tilde{x}$ and $x_{s}$ the corresponding subset for the original instance $x$.\footnote{In case of using our benchmarking tool with non-synthetic data we provide the possibility to provide a custom baseline. As default, baselining is done uniformly.}
    \begin{equation}
    \scriptsize
        Faith(f,E,x) =corr(\sum_{s \in S}E_{f} \left( x_{x_{s}=\tilde{x}_{s}}), f(x)-f(x_{x_{s}=\tilde{x}_{s}}) \right)
    \end{equation}
\subsection{Complexity}\label{sec:Comp}
Complexity \cite{bhatt2020evaluating} measures the number of features used in an explanation with a  fractional contribution distribution $\mathbb{P}_{g}$: the fractional contribution of feature $x_{i}$ to the total magnitude of the attribution: $  \mathbb{P}_{g}(i)= \frac{E_{f}(x)_{i}}{\sum |E_{f}(x)|} ; \mathbb{P}_{g} \in \{ \mathbb{P}_{g}(1) \ldots, \mathbb{P}_{g}(d)\}  $. The maximum value of complexity is $log(|E_{f}(x)|)$, where $|.|$  denotes the vector length. 
\begin{equation}
\scriptsize
    cpx(f;E;x)= -\sum_{i=1}^ {d} \mathbb{P}_{g}(i) ln(\mathbb{P}_{g}(i))
\end{equation}

\subsection{Reliability}\label{sec:reli}
Explanation methods should distinguish important from unimportant features at each time step and note changes over time. “Major” parts of an explanation should lie inside the ground truth mask $GT(x)$. XTSC-Bench includes the ground truth based measures relevance rank accuracy and relevance mask accuracy from \cite{arras2022clevr}. 

\begin{itemize}
    %
    \item Relevance Rank Accuracy \cite{arras2022clevr}: The relevance rank accuracy measures how much of the high intensity relevance lies within the ground truth.  We sort the top $K$ values of $E_{f}(x)$ in decreasing order $X_{top K }=\{ x_{1},..., x_{k}| E_{f}(x)_{1}>...>E_{f}(x)_{K}\}$. 
    \begin{equation}
   \scriptsize
        RACC=\frac{|X_{top K} \cap GT(x)|}{|GT(x)|}
    \end{equation}
    \item Relevance Mass Accuracy \cite{arras2022clevr}: The relevance mass accuracy is computed as the ratio of the sum of the Explanation values lying within the ground truth mask over the sum of all values.  
    \begin{equation}
    \scriptsize
        MACC=\frac{\sum_{E_{f}(x)_{i}\in GT(x)} E_{f}(x)_{i}}{\sum E_{f}(x)}
    \end{equation}
\end{itemize}
\section{Empirical Evaluation}\label{sec:Emp}
\begin{figure*}
    \centering
    \begin{subfigure}{0.45\textwidth}
    \includegraphics [width=0.9\columnwidth]{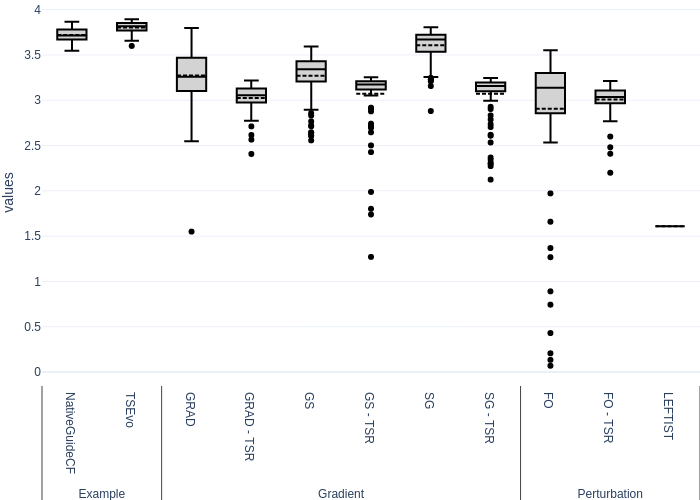}
    \caption{Complexity Univariate}\label{fig:ComplexityUni}        
    \end{subfigure}
    \begin{subfigure}{0.45\textwidth}
     \includegraphics [width=0.9\columnwidth]{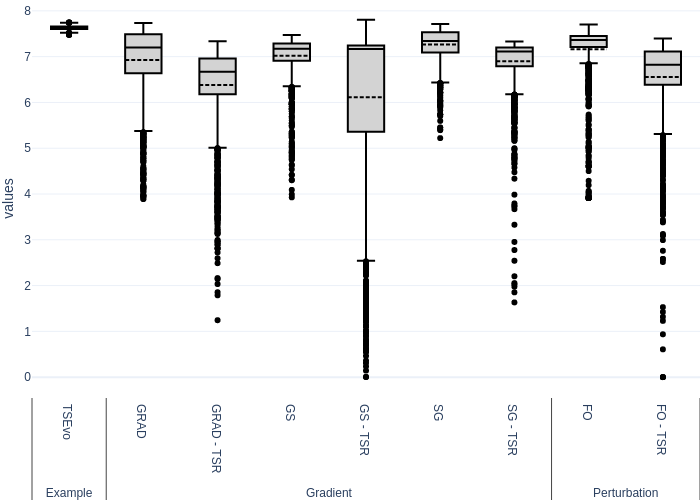}
    \caption{Complexity Multivariate}\label{fig:ComplexityMulti}       
    \end{subfigure}
    \\
    \begin{subfigure}{0.45\textwidth}
    \includegraphics [width=0.9\columnwidth]{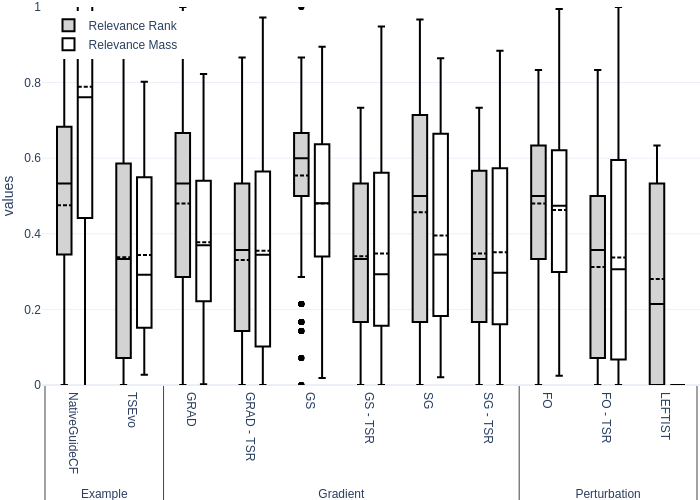}
    \caption{Reliability Univariate} \label{fig:ReliabilityUni}          
    \end{subfigure}
    \begin{subfigure}{0.45\textwidth}
    \includegraphics [width=0.9\columnwidth]{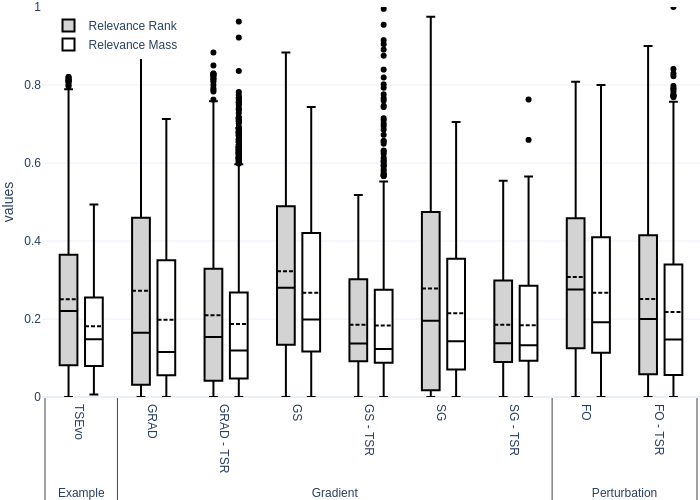}
    \caption{Reliability Multivariate} \label{fig:ReliabilityMulti}           
    \end{subfigure}
     \begin{subfigure}{0.45\textwidth}
    \includegraphics [width=0.9\columnwidth]{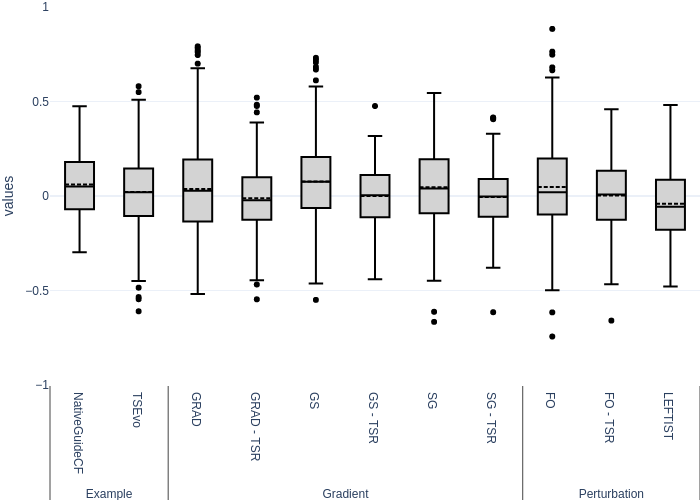}
    \caption{Faithfulness Univariate} \label{fig:FaithfulnessUni}          
    \end{subfigure}
     \begin{subfigure}{0.4\textwidth}
    \includegraphics [width=0.9\columnwidth]{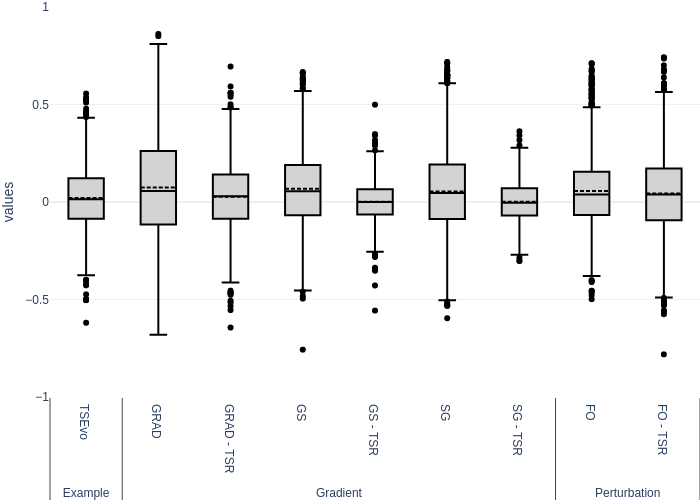}
    \caption{Faithfulness Multivariate}   \label{fig:FaithfulnessMulti}   
    \end{subfigure}
     \begin{subfigure}{0.45\textwidth}
    \includegraphics [width=0.9\columnwidth]{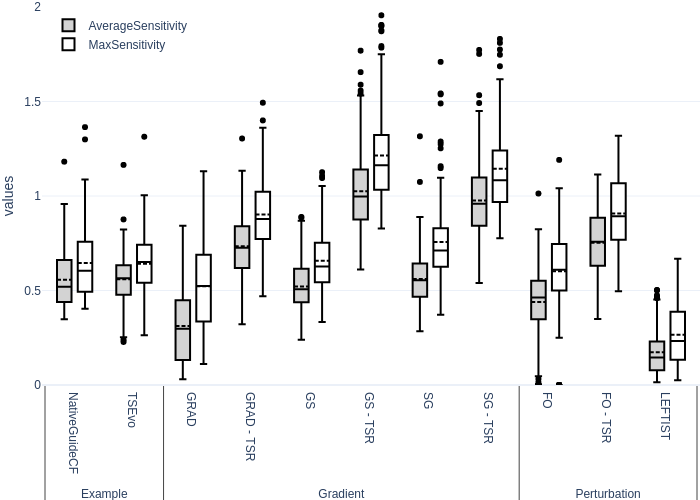}
    \caption{Robustness Univariate}\label{fig:RobustnessUni}           
    \end{subfigure}
     \begin{subfigure}{0.45\textwidth}
    \includegraphics [width=0.9\columnwidth]{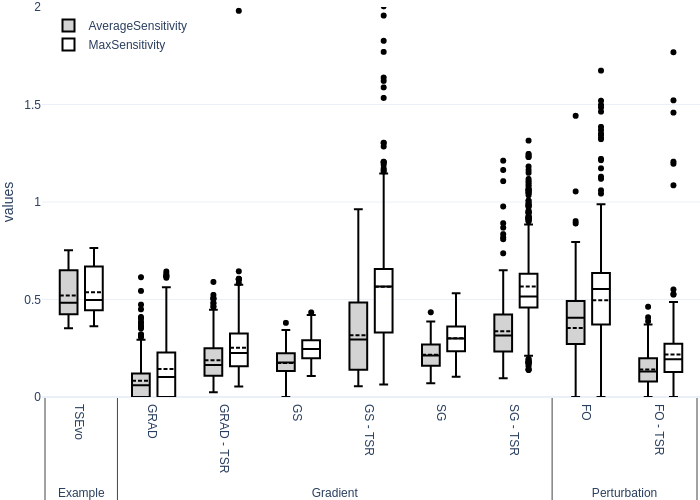}
    \caption{Robustness Multivariate} \label{fig:RoboustnessMulti}       
    \end{subfigure}

\caption{Explainer Performance on complexity, reliability, faithfulness, and robustness averaged over all datasets. The line denotes the median and the dotted line the mean. The start and end of the boxes are the first and third quartiles. 
Note, that Native Guide and LEFTIST only apply to univariate data and are therefore missing in the multivariate evaluation.}
\label{fig:Results}
\end{figure*}
This section compares the performance of 6 gradient- with 3 perturbation-based feature attribution methods and 2 example-based methods across Recurrent Neural Networks and Temporal Convolutional Networks for both the multi- and univariate synthetic time series (\Cref{sec:Synth}). The results are reported on a before unseen test set.
As gradient-based methods, we include Saliency (GRAD) \cite{simonyan2013deep}, Gradient Shap (GS) \cite{lundberg2017unified}, and Smooth Gradient (SG) \cite{smilkov2017smoothgrad} with and without Temporal Saliency Rescaling (TSR)  \cite{ismail2020benchmarking}. As perturbation-based, we include Feature Occlusion (FO) \cite{zeiler2014visualizing}  with and without Temporal Saliency Rescaling (TSR) \cite{ismail2020benchmarking} and LEFTIST \cite{guilleme2019agnostic}, an approach based on Lime adapted to time series.  
TSEvo \cite{hollig2022tsevo}, 
and Native Guide (NG) \cite{delaney2021instance} represent the example-based methods. For all methods, we use the implementation in TSInterpret \cite{hollig2023tsinterpret}. By employing XTSC-Bench, we evaluate the explainers'  capabilities on complexity, reliability, robustness, and faithfulness for all classifiers with an accuracy of over 90\%\footnote{Explainers are usually used to validate the inner-workings of well-performing classifiers. Classifiers with low accuracy cannot be expected to learn sufficient features to ensure an explainers reliability and classifier consistency.}. Additional information regarding the setting and the results can be found in our GitHub\footnote{\url{https://github.com/JHoelli/XTSC-Bench}}. 
\par
\Cref{fig:Results} summarizes the explainer-wise results on complexity, reliability, faithfulness, and robustness, averaged over all datasets and classifier models.
On complexity (\Cref{fig:ComplexityUni} and \Cref{fig:ComplexityMulti}), gradient- and perturbation-based methods provide less complex explanations than example-based methods. The results obtained by TSR contain slightly fewer attributions than the plain gradient- and perturbation-based methods, indicating that the explanations obtained after Temporal Saliency Rescaling are slightly easier to grasp. 
Averaging the obtained relevance scores on both the feature and time domain with TSR leads to a complexity decrease by eliminating areas with less relevance (e.g., single and small relevance scores on certain time steps). 
\par
The reliability (\Cref{fig:ReliabilityUni} and \Cref{fig:ReliabilityMulti}) on univariate data is higher than on multivariate data showing a decreasing capability of centering the explanation around the, in this case, known ground truth of all explainers with increasing data complexity. The on average lower relevance mask than rank indicates that while relevant features are found, the contribution of the found informative features to the overall relevance is low. Interestingly for both dataset types, the plain gradient- and perturbation-based methods (without TSR) perform slightly better on the Relevance Rank. On Relevance Mass, the difference between TSR and the plain approaches diverge (e.g., on univariate GRAD, TSR results in an improvement, on univariate GS, TSR results in a deterioration). 
\par
The faithfulness (\Cref{fig:FaithfulnessUni} and \Cref{fig:FaithfulnessMulti}) of the explainers to the classification models' behavior is similar for most explainers on the uni- and multivariate data. Least faithful is LEFTIST, as LEFTIST is the only approach relying on a local surrogate model instead of frequent classifier calls or the classifiers' inner workings (i.e., gradients).
\par
The results on robustness (\Cref{fig:RobustnessUni} and \Cref{fig:RoboustnessMulti}) indicate that on univariate data, perturbation-based approaches are less sensitive to small changes than example-based approaches. This results from perturbation-based approaches only relying on the perturbation function (which is constantly the same) and the classification model's output, while gradient-based approaches rely on a model's inner workings that possibly change with varying the input. 
\par 
Summarizing the results, no clear indication can be given on which explanation approaches should be preferred. No approach was able to dominate the plain gradient, and perturbation-based methods, which are included as baselines.
Both, traditional and time-series specific explainers show potential for improvement in all aspects. With increasing data complexity (univariate vs. multivariate), the metric performances diverge further, indicating a need for less complex, more reliable, and robust explainers, especially for multivariate time series classification. 

\section{Conclusion}
In this work, we propose XTSC-Bench, a benchmarking tool for the standardized evaluation of explainers for time series classifiers. XTSC-Bench aims to dissolve existing ambiguities and enable more comparability by providing synthetic datasets with informative features, from analogies to anomaly detection to moving features, trained models for the synthetic data, and options to evaluate custom data. A first empirical evaluation of the explainers implemented in TSInterpret \cite{hollig2023tsinterpret} showed that the current time series explainers leave potential for improvement, especially in providing reliable explanations for multivariate TSC. 
\bibliographystyle{IEEEtran}
\bibliography{paper}

\newpage
\appendix
This appendix provides additional explanations of the evaluation settings and visualizations of the results from \Cref{sec:Emp}. \Cref{app:Synth} provides insights into the generation process of the synthetic datasets, \Cref{sec:ExpApp} elucidates the used explainers, \Cref{app:counterman} explains tweaks to counterfactual explanation to use the benchmarking tool and \Cref{app:info} - \Cref{app:orig} visualize additional results.
\subsection{Data Generation} \label{app:Synth}
The synthetic datasets described in \Cref{sec:Synth} were generated as described in Ismail et al. \cite{ismail2020benchmarking}\footnote{\url{https://github.com/ayaabdelsalam91/TS-Interpretability-Benchmark}}. XTSC-Bench provides the generated data for 50 time steps with a feature size of 1 and 50. The data is generated based on 6 time processes with $\epsilon_{t} \sim N(0,1)$:
\begin{itemize}
   \item Gaussian ($\mu = 0 ,  \sigma = 0 $): \\ $X_{t}= \epsilon_{t}$
   \item Harmonic: \\ $X(t)= sin(2\pi 2t) + e_{t}$ 
   \item Pseudo Periodic ($A_{t} \sim N(0,0.5)$, $f_{t} \sim N(2,0.01)$): \\
   $X(t)=A_{t}sin(2\pi f_{t}, t)+ \epsilon_{t}$
   \item Autoregressive ($p=1$, $\varphi=0.9$): \\ $X_{t}= \sum_{i=1}^{p} \varphi X_{t-i}+\epsilon_{t}$
   \item Continuous Autoregressive ($\varphi=0.9$, $\sigma=0.1$) : \\
   $X_{t}= \varphi X_{t-1}+\sigma(1-\varphi)^2 * \epsilon + \epsilon_{t}$
   \item NARMA ($n=10$, $U \sim U(0.05)$): 
   \\
   $X_{t}=0.3X_{t-1}+0.05X_{t-1} \sum_{i=0}^{n-1}X_{t-1}+1.5U(t-(n-1))*U(t) +0.1+\epsilon_{t}$
\end{itemize}
The obtained datasets highlight predefined informative features by adding a constant to the positive class or subtracting a constant for negative classes. As visualized in \Cref{fig:InformFeat}, the informative features can take various forms to replicate different ground truths:
\begin{itemize}
    \item \textbf{Middle vs. Moving vs. Positional}: denotes the location of the informative features.
    \item \textbf{Small vs. Normal vs. Rare Time / Feature}: refers to the size (number) of informative features. For Normal, more than 35\% of all features are informative. For Small, less than 10\% of all features are informative. A time or feature is rare if less than 5\% of all features are informative.
\end{itemize}
Overall we obtain 60 univariate datasets and 60 multivariate datasets (6 time process $\times$ 10 informative features).
\par
\emph{Practical Note}: The function evaluate\_synthetic allows filtering the synthetic datasets by providing the variable \emph{types}, enabling the evaluation of designated informative features and time series processes. For example, providing $types = ['Rare']$ would allow the evaluation of explainers for anomaly detection.
\subsection{Explanation Approaches}\label{sec:ExpApp}
According to the taxonomy provided by Höllig et al. \cite{hollig2023tsinterpret},  we divided the explanation approaches into gradient-based and perturbation-based feature attribution methods and example-based approaches. Gradient-based feature attribution methods assign a relevance score to a machine learning model's inputs based on the classifier model's gradients. Perturbation-based feature attribution methods also assign relevance scores, however, they obtain the relevance scores by observing the classifier's output while masking parts of the input. In contrast to feature attribution methods, example-based methods return a manipulated version of the input instance $x$, e.g., to show how a counterexample looks. We evaluate all the explainers applicable to uni- and multivariate time series implemented in TSInterpret \cite{hollig2023tsinterpret}. 
\begin{itemize}
    \item \textbf{TSR}: Temporal Saliency Rescaling (TSR) is a wrapper for well-known perturbation (e.g., Feature Occlusion (FO) \cite{zeiler2014visualizing}) and gradient-based Feature Attribution Methods (e.g., Gradient Shap (GS) \cite{lundberg2017unified}, Integrated Gradient (IG) \cite{sundararajan2017axiomatic}, Saliency (GRAD) \cite{simonyan2013deep}, Smooth Gradients (SG) \cite{smilkov2017smoothgrad})  developed by Ismail et al. \cite{ismail2020benchmarking}. TSR is applied after the explanation calculation and decouples the time and feature domain by computing time and feature relevance scores.
    \item \textbf{LEFTIST}: LEFTIST \cite{guilleme2019agnostic} adapts SHAP \cite{lundberg2017unified} and LIME \cite{ribeiro2016should} to time series. An interpretable model is fitted locally by perturbing the input x meaningfully by segmenting the original time series into interpretable components and perturbing those components with a) linear interpolation, b) a constant, or c) a background obtained from a reference set. In this work, we make use of variant c) and segmentation of size $10$.
    \item \textbf{TSEvo}: TSEvo \cite{hollig2022tsevo} generated counterfactuals for uni-and multivariate time series using time series specific perturbation functions (e.g., perturbing the frequency domain). We use the authentic information transformer and run TSEvo for 100 epochs.
    \item \textbf{NG}: Delaney et al. \cite{delaney2021instance} propose using a Native Guide (i.e., an existing instance in the data that is the nearest unlike neighbor to the original instance) to generate counterfactuals. The original time series is thereby manipulated with the Native Guide by replacing the most important features of $x$ (obtained with, e.g., GradCam \cite{selvaraju2017grad}) with the Native Guide. 
\end{itemize}
\subsection{Feature Ranking and Mass Calculation for example-based Explainers}\label{app:counterman}
Replacing features to calculate the robustness (\Cref{sec:Roboust}), reliability (\Cref{sec:reli}), or faithfulness metrics (\Cref{sec:Faith}) relies on either ranking the most important features or calculating relevance masks. For feature attribution methods, this is straightforward, as feature attribution methods return relevance scores.  However, example-based methods return a manipulated version of the original inputs. The changes made to the original input can usually not be directly interpreted as relevance scores. To be able to still use the metrics with example-based methods, we calculate the fraction of change $\Delta x = \frac{(x - E_{f}(x))}{x}$.\par
\emph{Practical Note}: The synthetic data is normalized to zero and one. For non-synthetic data, a feature range needs to be provided.
\subsection{Results split on Informative Features  Types}\label{app:info}
\begin{figure}
    \centering
    \begin{subfigure}{0.45\textwidth}
    \includegraphics [width=0.9\columnwidth]{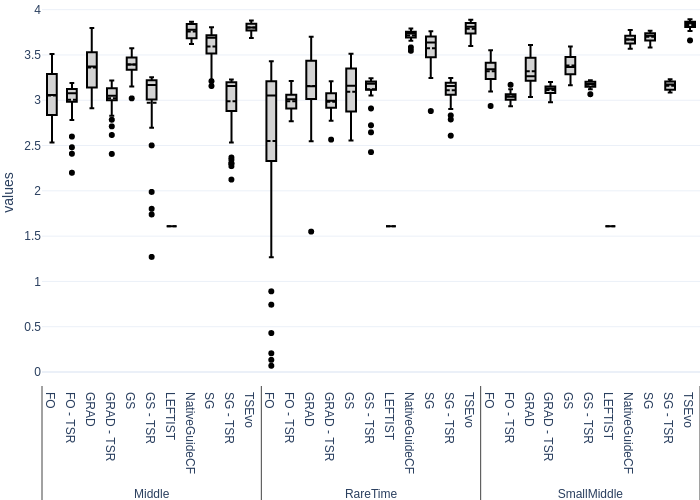}
    \caption{Complexity Univariate}        
    \end{subfigure}
    \\
    \begin{subfigure}{0.45\textwidth}
    \includegraphics[width=0.9\columnwidth]{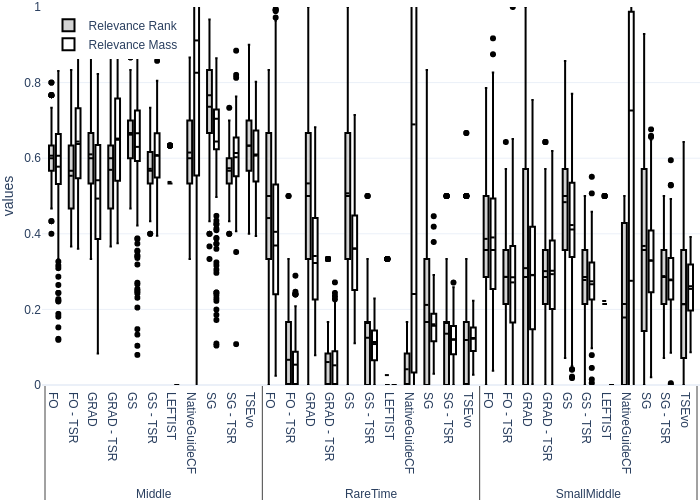}
    \caption{Reliability Univariate}        
    \end{subfigure}
    \\
    \begin{subfigure}{0.45\textwidth}
    \includegraphics [width=0.9\columnwidth]{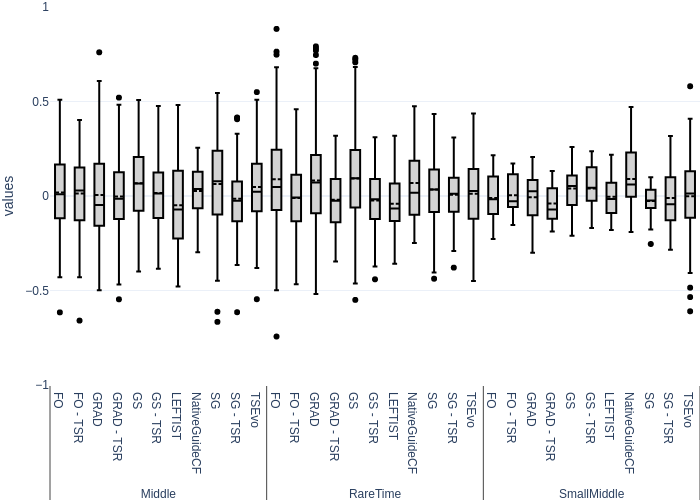}
    \caption{Faithfulness Univariate}        
    \end{subfigure}
    \\
     \begin{subfigure}{0.45\textwidth}
    \includegraphics [width=0.9\columnwidth]{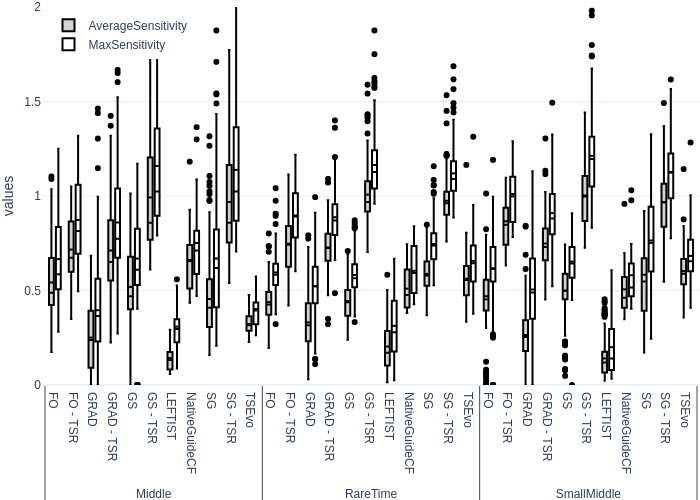}
    \caption{Robustness Univariate}        
    \end{subfigure}

\caption{Informative-feature-wise explainer performance on complexity, reliability, faithfulness, and robustness averaged over all generation processes. }
\label{fig:info_feat}
\end{figure}

\begin{figure*}
    \centering
    \begin{subfigure}{\textwidth}
    \includegraphics [width=0.9\textwidth]{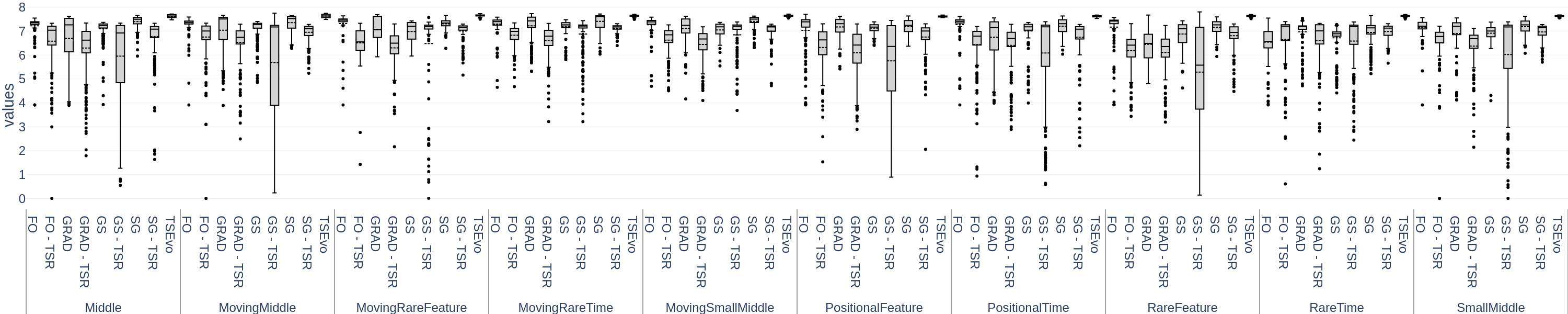}
    \caption{Complexity Multivariate}        
    \end{subfigure}

    \begin{subfigure}{\textwidth}
    \includegraphics[width=0.9\textwidth]{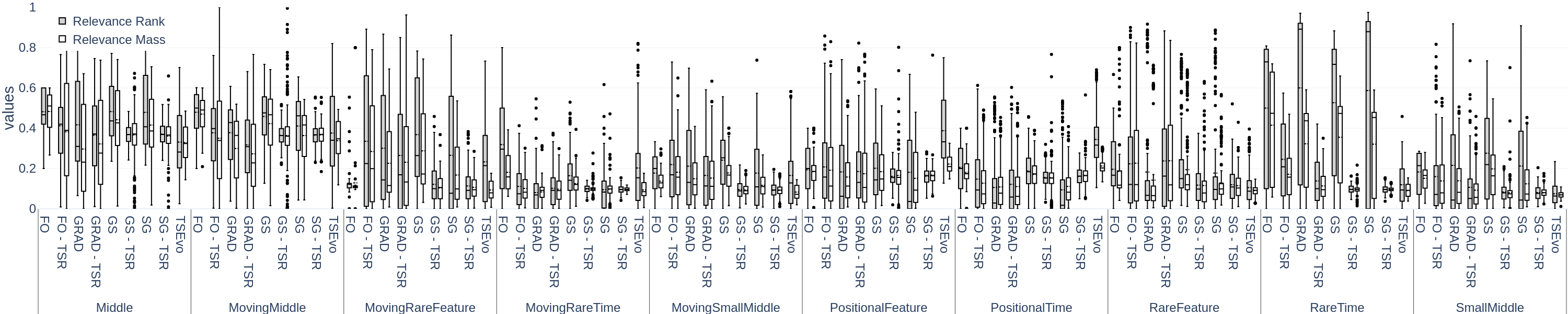}
    \caption{Reliability Multivariate}        
    \end{subfigure}

     \begin{subfigure}{\textwidth}
    \includegraphics [width=0.9\textwidth]{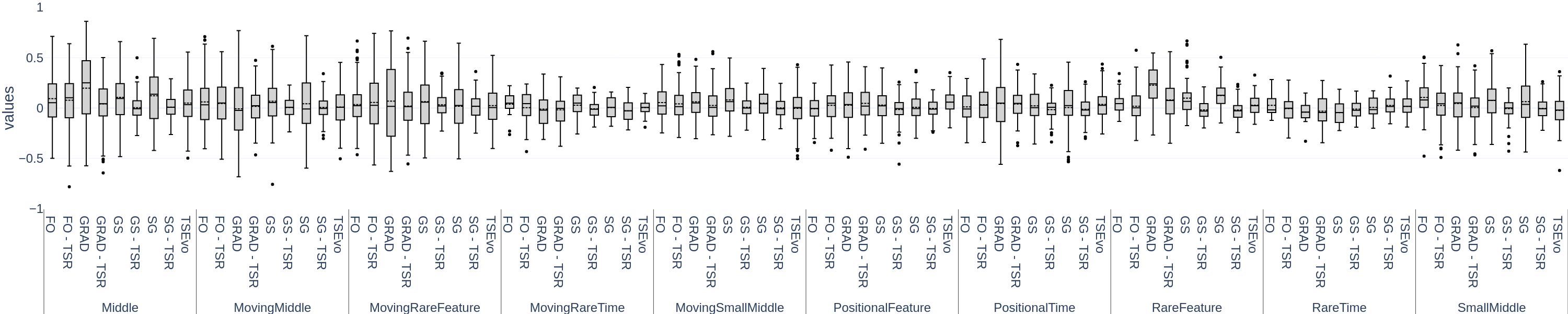}
    \caption{Faithfulness Multivariate}        
    \end{subfigure}

     \begin{subfigure}{\textwidth}
    \includegraphics[width=0.9\textwidth]{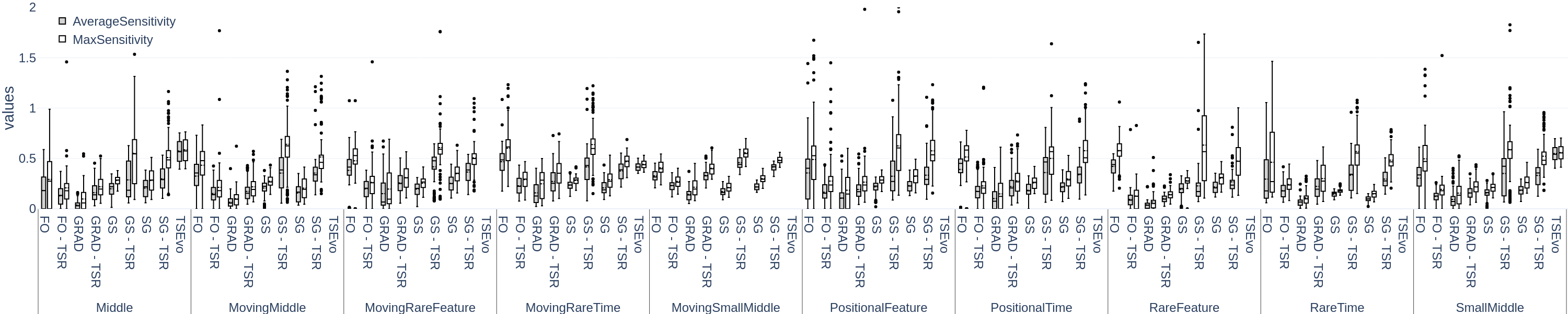}
    \caption{Robustness Multivariate}        
    \end{subfigure}

\caption{Informative-feature-wise explainer performance on complexity, reliability, faithfulness, and robustness averaged over all generation processes.}
\label{fig:info_feat_multi}
\end{figure*}

\Cref{fig:info_feat} and \Cref{fig:info_feat_multi} show the results averaged over all time series processes split on the explainer and the informative feature type.  Due to the availability of only one feature, all feature-based datasets are missing for univariate data.\par
The explainers perform similarly on robustness, faithfulness, and complexity on uni- and multivariate data across the different informative feature types. On reliability, the informative feature type has on univariate time series a huge impact on the performance of all explainers. The reliability on univariate and multivariate data is the largest for all explainers on the informative feature 'Middle' (over 30\% of all time steps and features are informative) and decreases with the number of informative features ('Middle' $\rightarrow$ 'SmallMiddle' $\rightarrow$ 'Rare').

\subsection{Results split on Classifier Models}
\begin{figure*}
    \centering
    \begin{subfigure}{0.45\textwidth}
    \includegraphics [width=0.9\columnwidth]{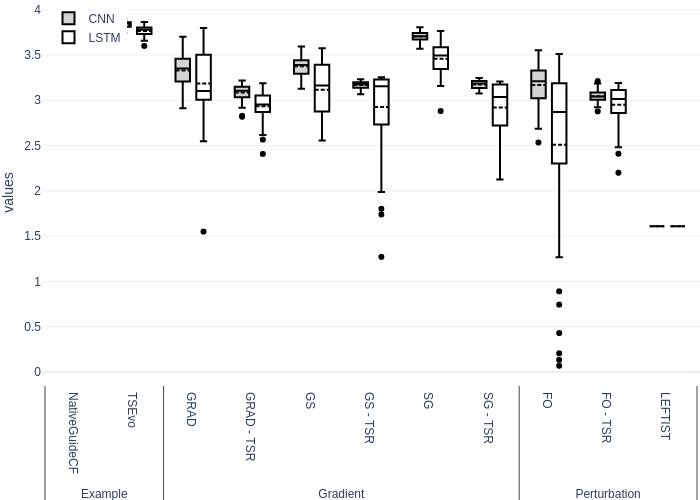}
    \caption{Complexity Univariate}        
    \end{subfigure}
    \begin{subfigure}{0.45\textwidth}
    \includegraphics [width=0.9\columnwidth]{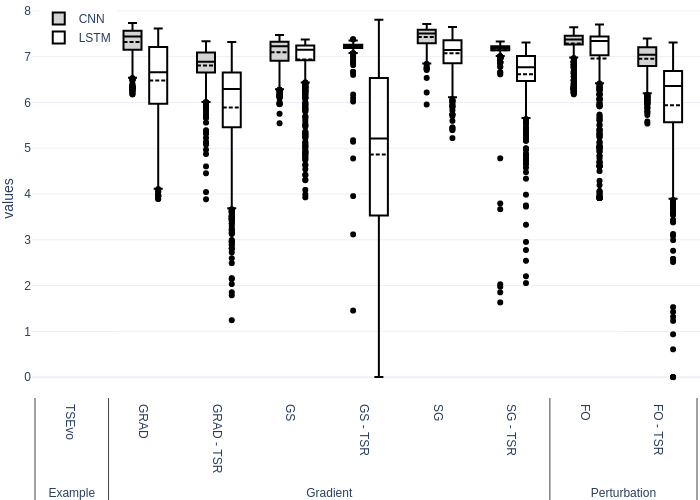}
    \caption{Complexity Multivariate}        
    \end{subfigure}
    \\
    \begin{subfigure}{0.45\textwidth}
    \includegraphics[width=0.9\columnwidth]{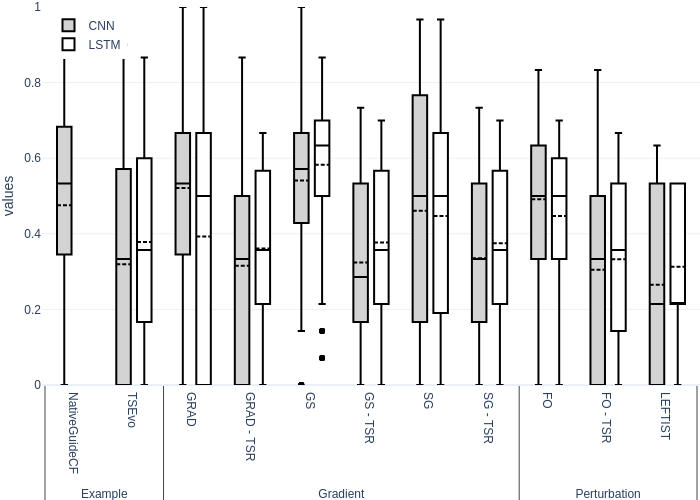}
    \caption{Reliability Univariate}        
    \end{subfigure}
    \begin{subfigure}{0.45\textwidth}
    \includegraphics[width=0.9\columnwidth]{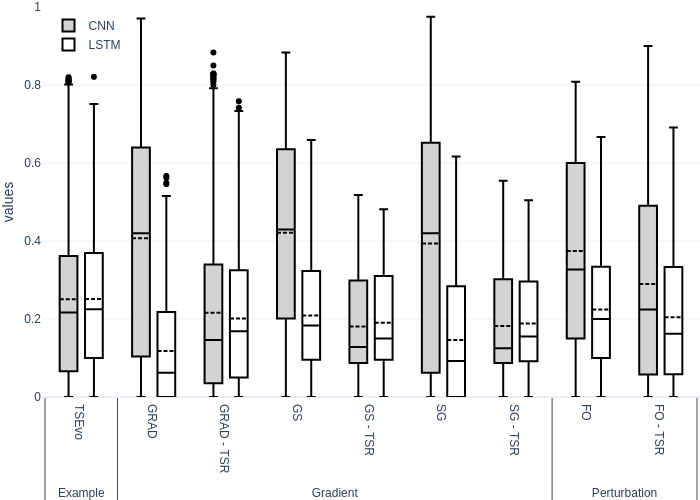}
    \caption{Reliability Multivariate}        
    \end{subfigure}
     \begin{subfigure}{0.45\textwidth}
    \includegraphics [width=0.9\columnwidth]{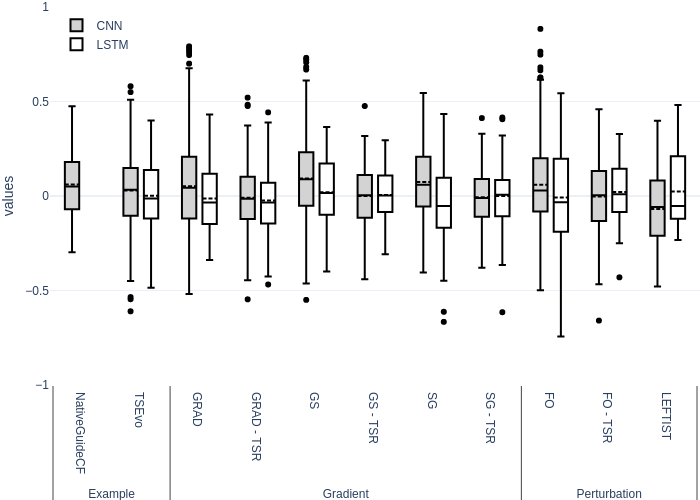}
    \caption{Faithfulness Univariate}        
    \end{subfigure}
     \begin{subfigure}{0.45\textwidth}
    \includegraphics [width=0.9\columnwidth]{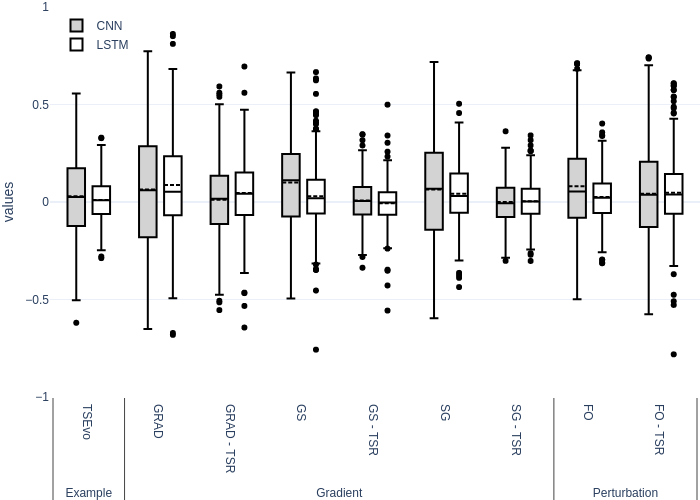}
    \caption{Faithfulness Multivariate}        
    \end{subfigure}
     \begin{subfigure}{0.45\textwidth}
    \includegraphics [width=0.9\columnwidth]{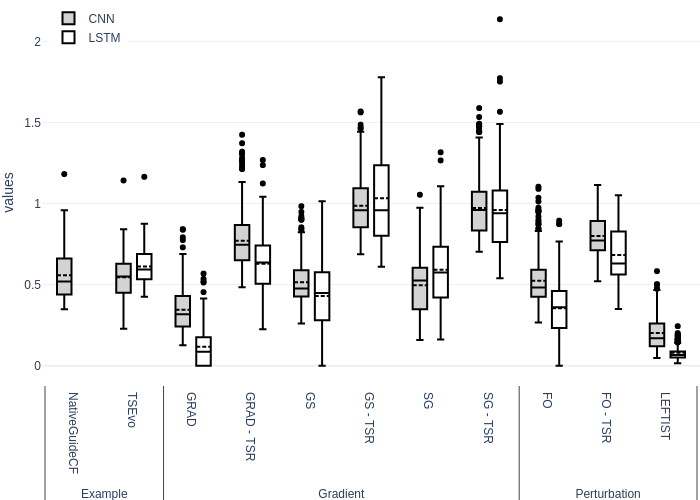}
    \caption{Robustness Univariate}        
    \end{subfigure}
     \begin{subfigure}{0.45\textwidth}
    \includegraphics[width=0.9\columnwidth]{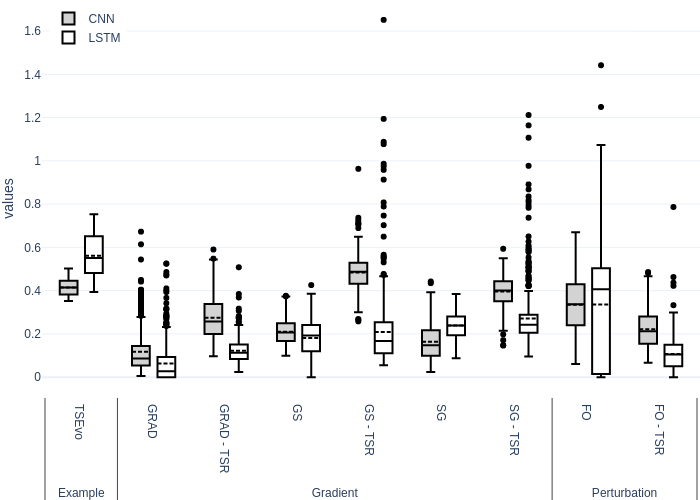}
    \caption{Robustness Multivariate}        
    \end{subfigure}

\caption{Explainer Performance on complexity, reliability, faithfulness, and robustness averaged over all datasets and split on the used classifier.}
\label{fig:models}
\end{figure*}

\Cref{fig:models} shows the complexity, reliability, robustness,  and faithfulness averaged over all datasets and split on the classification model to be explained. If no box for a model is provided, either the model's accuracy was below 90\% or the explainer was not applicable to the classifier. \par
On average, explainers on LSTMs result in less complex explanations than CNN. Explainers on CNN and LSTM perform similarly on reliability for example and perturbation-based approaches. On faithfulness and robustness, explainers on CNNs and LSTMs show no dominant behavior. Therefore, on most metrics, the classifier type has no larger influence on the explainer's performance. However, on multivariate data, a slightly higher reliability can be observed for gradient-based approaches without TSR  for explanations based on CNN. The performance increase cannot be observed after applying TSR, indicating that 'traditional' gradient-based approaches work well for multivariate data with CNN Classifiers and that LSTM-based gradient explainers need improvement.

\subsection{Faithfulness: Comparison of Baselines}\label{app:orig}
\begin{figure}[H]
    \centering
    \includegraphics[width=0.7\columnwidth]{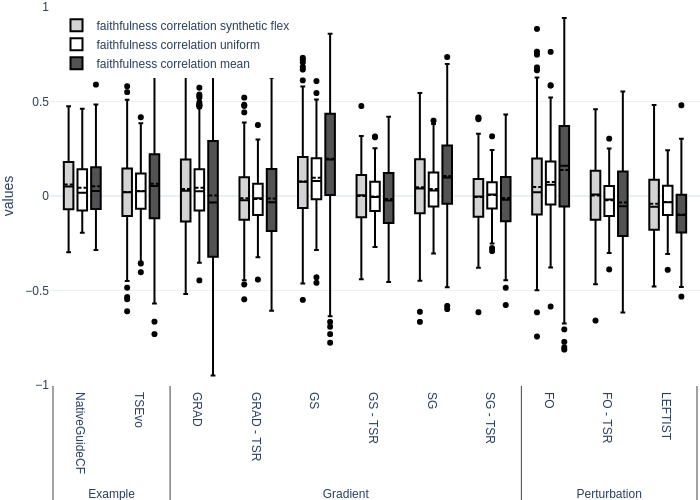}
    \caption{Comparison of baselines used in the calculation of the faithfulness metric.}
    \label{fig:Faith}
\end{figure}

\Cref{fig:Faith} compares the faithfulness metric with the generation baseline used for the synthetic data to the 'traditionally' used baselines mean and uniform. As the uniform baseline performs, on average, similar to the known generation process baseline, we advise users with non-synthetic data to use the uniform baseline.

\end{document}